\documentclass[10pt,twocolumn,letterpaper]{article}

\usepackage{iccv}
\usepackage{times}
\usepackage{epsfig}
\usepackage{graphicx}
\usepackage{amsmath}
\usepackage{amssymb}

\usepackage{subfigure}

\usepackage{interval}

\usepackage{caption}

\usepackage{pifont}

\usepackage{makecell}
\usepackage{booktabs}
\usepackage{multirow}
\usepackage{siunitx}

\usepackage[T1]{fontenc}
\usepackage{fix-cm}

\newcolumntype{?}{!{\vrule width 1pt}}
\newcolumntype{C}[1]{>{\centering}m{#1}}
\newcolumntype{F}[1]{>{\centering\arraybackslash}p{#1}}

\newcolumntype{X}{@{\hskip\tabcolsep\vrule width 1.5pt\hskip\tabcolsep}}

\newcommand{\myfigurethreecol}[1]{
\begin{minipage}[b]{.14\textwidth}
\includegraphics[width=1.09\linewidth]{#1}
\end{minipage}
}

\newcommand{\myfigurethreecolcaption}[2]{
\begin{minipage}[b]{.14\textwidth}
\includegraphics[width=1.09\linewidth]{#1}
\caption{{\small {#2}}}
\end{minipage}
}

\newcommand{\myfiguresixcol}[1]{
\begin{minipage}[b]{.14\textwidth}
\includegraphics[width=1.1\linewidth]{#1}
\end{minipage}
}

\newcommand{\myfiguresixcolcaption}[2]{
\begin{minipage}[b]{.14\textwidth}
\includegraphics[width=1.1\linewidth]{#1}
\caption{{\small {#2}}}
\end{minipage}
}

\usepackage[pagebackref=true,breaklinks=true,letterpaper=true,colorlinks,bookmarks=false]{hyperref}

 \iccvfinalcopy 

\def\iccvPaperID{3} 

\ificcvfinal\fi
\begin{document}

\title{Using Cross-Model EgoSupervision \\ to Learn Cooperative Basketball Intention}

\author{Gedas Bertasius\\
University of Pennsylvania\\
{\tt\small gberta@seas.upenn.edu}
\and
Jianbo Shi\\
University of Pennsylvania\\
{\tt\small jshi@seas.upenn.edu}
}

\maketitle

\begin{abstract}

We present a first-person method for cooperative basketball intention prediction: we predict with whom the camera wearer will cooperate in the near future from unlabeled first-person images. This is a challenging task that requires inferring the camera wearer's visual attention, and decoding the social cues of other players. Our key observation is that a first-person view provides strong cues to infer the camera wearer's momentary visual attention, and his/her intentions. We exploit this observation by proposing a new cross-model EgoSupervision learning scheme that allows us to predict with whom the camera wearer will cooperate in the near future, without using manually labeled intention labels. Our cross-model EgoSupervision operates by transforming the outputs of a pretrained pose-estimation network, into pseudo ground truth labels, which are then used as a supervisory signal to train a new network for a cooperative intention task. We evaluate our method, and show that it achieves similar or even better accuracy than the fully supervised methods do.

\end{abstract}

\vspace{-0.5cm}


\section{Introduction}

Consider a dynamic scene such as Figure~\ref{task_fig}, where you, as the camera wearer, are playing basketball.  You need to make a decision with whom you will cooperate to maximize the overall benefit for your team. Looking ahead at your teammates, you make a conscious decision and then 2-3 seconds afterwards you perform a cooperative action such as passing the ball.  

In a team sport such as basketball, an effective cooperation among teammates is essential. Thus, in this paper, we aim to investigate whether we can use a single first-person image to infer with whom the camera wearer will cooperate 2-3 seconds from now? This is a challenging task because predicting camera wearer's cooperative intention requires 1) inferring his/her momentary visual attention, 2) decoding dominant social signals expressed by other players who want to cooperate, and 3) knowing who your teammates are when the players are not wearing any team-specific uniforms.

\begin{figure}
\centering

\includegraphics[width=1\linewidth]{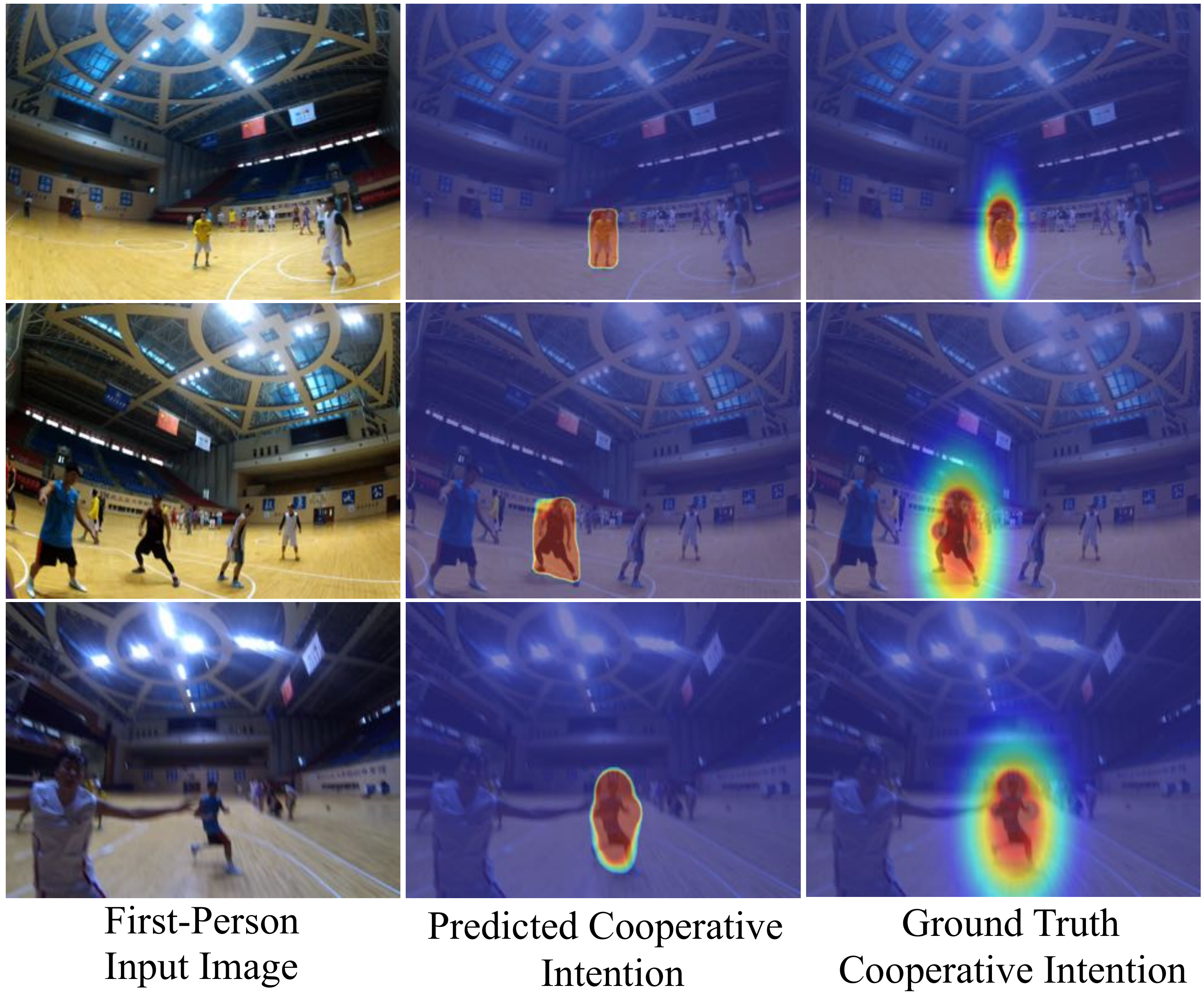}

\captionsetup{labelformat=default}
   \setcounter{figure}{0}
    \caption{With whom will I cooperate after 2-3 seconds? Given an \textbf{unlabeled} set of first-person basketball images, we predict with whom the camera wearer will cooperate 2 seconds from now. We refer to this problem as a cooperative basketball intention prediction.\vspace{-0.6cm}}
    \label{task_fig}
\end{figure}

\begin{figure*}
\begin{center}
   \includegraphics[width=0.9\linewidth]{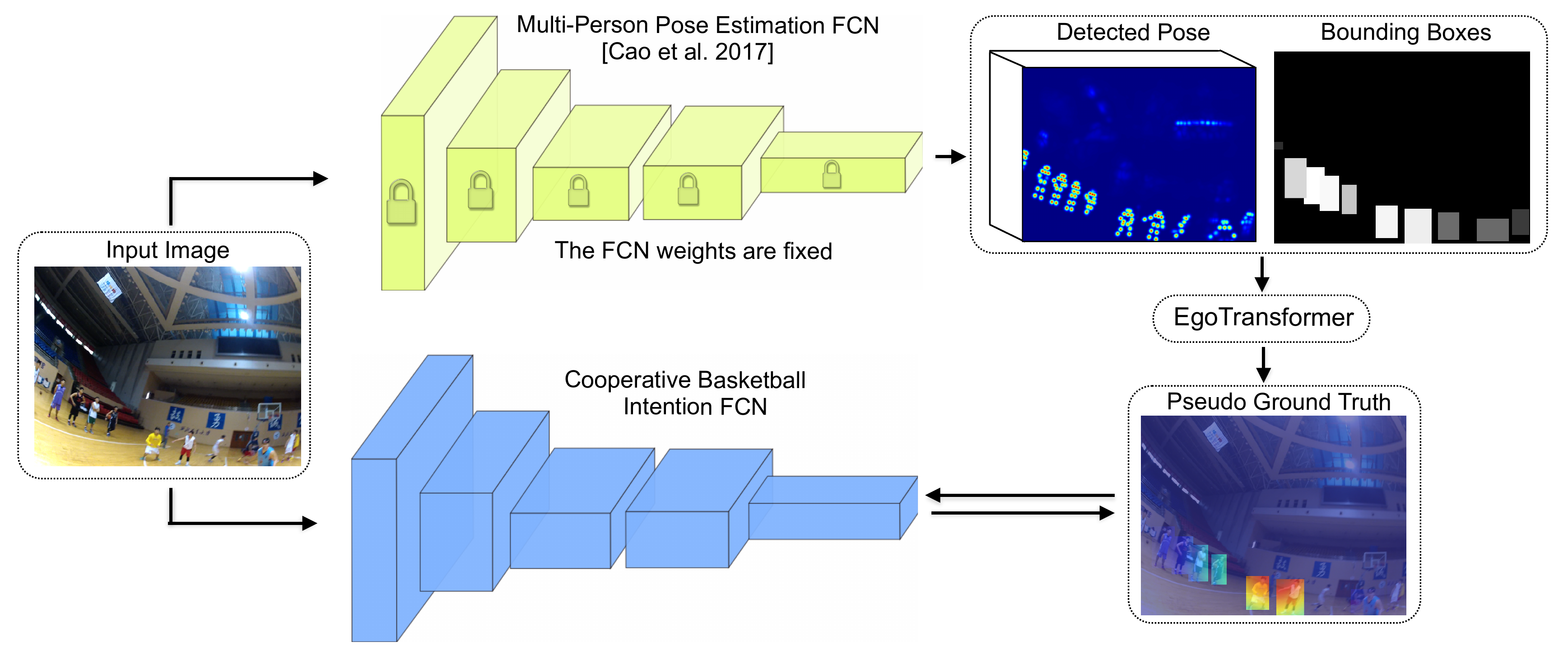}
\end{center}
\vspace{-0.4cm}
\caption{The illustration of our cross-model EgoSupervision training scheme. As our base model we use a multi-person pose estimation network from~\cite{DBLP:journals/corr/CaoSWS16}, which predicts  1) pose estimates of all people in a given first-person image and 2) the bounding boxes around each person. Next, we feed these outputs to an EgoTransformer, which transforms them such that the transformed output would approximately capture the camera wearer's attention and intentions. Then, we use such transformed output as a supervisory signal to train the network for our cooperative basketball intention task.\vspace{-0.5cm}}
\label{fig:train_arch}
\end{figure*}

To make this problem even more challenging we ask a question: ``Can we infer cooperative basketball intention without manually labeled first-person data?''. Building an unsupervised learning framework is important because manually collecting basketball intention labels is a costly and a time consuming process. In the context of a cooperative basketball intention task, an annotator needs to have highly specific basketball domain knowledge. Such a requirement limits the scalability of the annotation process because such annotators are difficult to find and costly to employ.

However, we conjecture that we can learn cooperative basketball intention in an unsupervised fashion by exploiting the signal provided by the first-person camera. What people see reflects how they are going to act. A first-person camera placed on a basketball player's head allows us to indirectly tap into that person's mind and reason about his/her internal state based on what the camera wearer sees. To do so we propose a novel cross-model EgoSupervision learning scheme, which allows us to learn the camera wearer's intention without the manually labeled intention data. Our cross-model EgoSupervision scheme works as follows. First we transform the output of a pretrained pose-estimation network such that it would approximately reflect the camera wearer's internal state such as his/her visual attention and intentions. Then, we use such transformed output as a supervisory signal to train another network for our cooperative basketball intention task. We show that such a learning scheme allows us to train our model without manually annotated intention labels, and achieve similar or even better results as the fully supervised methods do.

\section{Related Work}

\textbf{First-Person Vision.} In the past, most first-person methods have focused on first-person object detection~\cite{DBLP:journals/ijcv/LeeG15,BMVC.28.30,conf/cvpr/RenG10,conf/cvpr/FathiRR11,gberta_2017_RSS}, or activity recognition~\cite{Soran2015,Singh_2016_CVPR,PirsiavashR_CVPR_2012_1,Li_2015_CVPR,ma2016going,Fathi:2011:UEA:2355573.2356302}. Several methods have employed first-person videos to summarize videos ~\cite{DBLP:journals/ijcv/LeeG15,Lu:2013:SSE:2514950.2516026} while recently the work in~\cite{Su2016} proposed to predict the camera wearer's engagement detection from first-person videos. The work in~\cite{Fathi_socialinteractions:} used a group of people wearing first-person cameras to infer their social interactions such as monologues, dialogues, or discussions. The method in~\cite{park_force} predicted physical forces experienced by the camera wearer, while the work in~\cite{conf/cvpr/KitaniOSS11} recognized the activities performed in various extreme sports. Several recent methods~\cite{park_ego_future,park_cvpr:2017} also predicted the camera wearer's movement trajectories. Finally, first-person cameras have also been used for various robotics applications~\cite{Ryoo:2015:RAP:2696454.2696462,DBLP:journals/corr/GoriAR15}

In comparison to these prior methods, we propose a novel cooperative basketball intention prediction task, that allows us to study cooperative behaviors of the basketball players. Furthermore, we note that these prior first-person methods (except~\cite{conf/cvpr/KitaniOSS11}) rely on manually annotated labels for their respective tasks whether it would be an object-detection, activity recognition, intention prediction or some other task. Instead, in this work, we demonstrate that we can solve a challenging cooperative basketball intention prediction task without using annotated first-person intention labels, which are time consuming and costly to obtain. 




\textbf{Knowledge Transfer across Models.} With the introduction of supervised CNN models~\cite{NIPS2012_4824}, there has been a lot of interest in adapting generic set of features~\cite{Donahue_ICML2014} for different tasks at hand~\cite{NIPS2014_5418,gberta_2015_CVPR,girshick2014rcnn,DBLP:journals/corr/XieT15,ren2015faster,Sermanet_overfeat:integrated}. Recently, generic image classification features were successfully used for the tasks such as edge detection~\cite{gberta_2015_CVPR,DBLP:journals/corr/XieT15}, object detection~\cite{girshick2014rcnn,ren2015faster,Sermanet_overfeat:integrated}, and semantic segmentation~\cite{gberta_2016_CVPR,DBLP:journals/corr/LinSRH15,DBLP:journals/corr/LongSD14,DBLP:journals/corr/ChenPKMY14}. More related to our work, a recent line of research investigated how to transfer knowledge across different models by a combination of parameter updates~\cite{Aytar11,DuanICML2012,Hoffman_ICLR2013}, transformation learning~\cite{Kulis:2011:YSY:2191740.2191798,DBLP:conf/cvpr/GongSSG12}, network distillation~\cite{DBLP:journals/corr/HintonVD15} or cross-model supervision~\cite{Hoffman_2016_CVPR,Gupta_2016_CVPR}. The most similar to our work are the methods in~\cite{Hoffman_2016_CVPR,Gupta_2016_CVPR} that use cross-model supervision to transfer knowledge from one model to another.  

All of the above methods focus on the third-person data. In contrast, we show how to exploit a first-person view to solve a novel camera wearer's cooperative intention prediction task without using manually labeled first-person data.

\section{Learning Cooperative Basketball Intention}

The goal of our cooperative basketball intention task is to predict with whom the camera wearer will cooperate in the near future. Formally, we aim to learn a function $g(I_i)$ that takes a single first-person image $I_i$ as an input and outputs a per-pixel likelihood map, where each pixel indicates the cooperation probability. Ideally, we would want such function to produce high probability values at pixels around the person with whom the camera wearer will cooperate, and low probability values around all the other pixels. 

We implement $g(I_i)$ via a fully convolutional neural network based on the architecture of a multi-person pose estimation network in~\cite{DBLP:journals/corr/CaoSWS16}. Let $\hat{y}$ denote a per-pixel mask that is given to our network as a target label. We refer to $\hat{y}$ as a \textit{pseudo} ground truth because we obtain it automatically instead of relying on the manually annotated intention labels. Then, we learn our cooperative basketball intention model by optimizing the following cross-entropy loss objective:

\vspace{-0.4cm}
\begin{equation}
\begin{split}
\label{CE_loss_eq}
  \mathcal{L}^{(i)}= -\sum_{j=1}^{N} \hat{y}^{(i)}_j \log g_j(I_i) +(1-\hat{y}^{(i)}_j) \log \left(1-g_j(I_i)\right), 
\end{split}
\end{equation}

where $\hat{y}^{(i)}_j$ is the pseudo ground truth value of image $I_i$ at pixel $j$, $g_j(I_i)$ refers to our network's output at pixel $j$, and $N$ denotes the number of pixels in an image. We now explain how we obtain the pseudo ground truth data $\hat{y}$.




\subsection{EgoTransformer}

To construct a pseudo ground truth supervisory signal $\hat{y}$, we transform the output of a pretrained multi-person pose estimation network~\cite{DBLP:journals/corr/CaoSWS16}, such that it would approximately capture the camera wearer's internal state such as his/her visual attention, and intentions. We do so using our proposed EgoTransformer scheme.

Let $f(I_i)$ denote a pretrained fully convolutional network from~\cite{DBLP:journals/corr/CaoSWS16} that takes a first-person image as an input, and outputs the 1) pose part estimates of every person in an image, and 2) their bounding-box detections. We note that the pretrained network $f$ was never trained on any first-person images. Then, formally, let $B \in \mathbb{R}^{n \times 5}$ denote the bounding box of people detected by $f$. Each of $n$ detected bounding boxes is parameterized by $5$ numbers $(x,y,h,w,c)$ denoting the top-left bounding-box coordinates $(x,y)$, the height $h$, and width $w$ of the bounding box, and its confidence value $c$. Additionally, let $P \in \mathbb{R}^{n \times 18 \times 2}$ denote the predicted $(x,y)$ locations of $18$ pose parts (see~\cite{DBLP:journals/corr/CaoSWS16}) for each of $n$ detected people. 

Then our goal is to come up with a transformation function $T(B^{(i)},P^{(i)})$ that takes these two outputs and transforms them into a per-pixel pseudo ground truth mask $\hat{y}^{(i)}$ for our cooperative basketball intention prediction task.

We do so by exploiting three different characteristics encoded in a first-person view: 1) egocentric location prior, 2) egocentric size prior, and 3) egocentric pose prior. All of these characteristics can be used to reason about the camera wearer's internal state.

\captionsetup{labelformat=empty}
\captionsetup[figure]{skip=5pt}

\begin{figure}
\centering

\myfigurethreecol{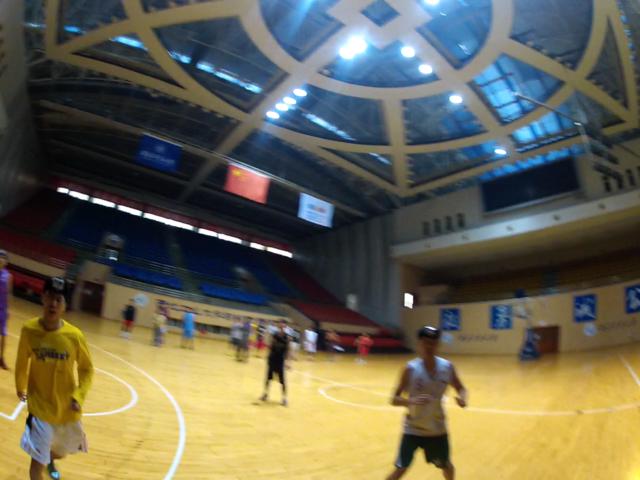}
\myfigurethreecol{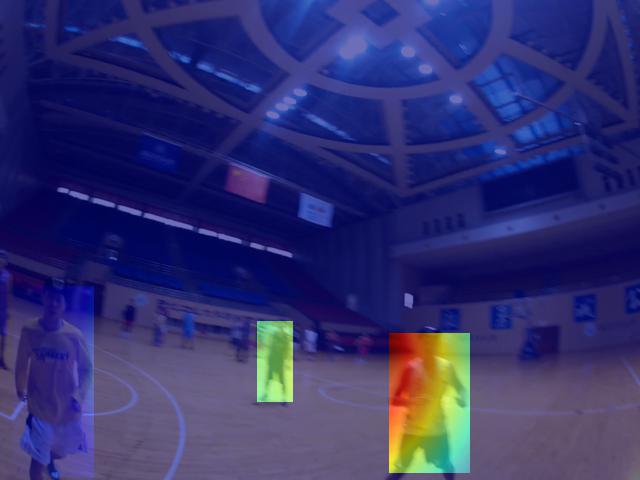}
\myfigurethreecol{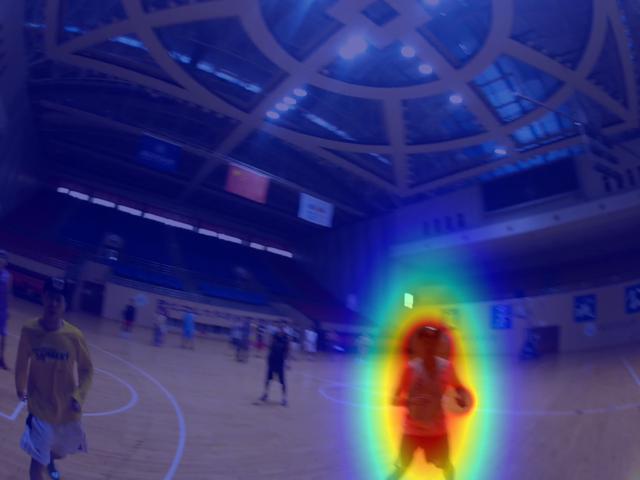}

\myfigurethreecol{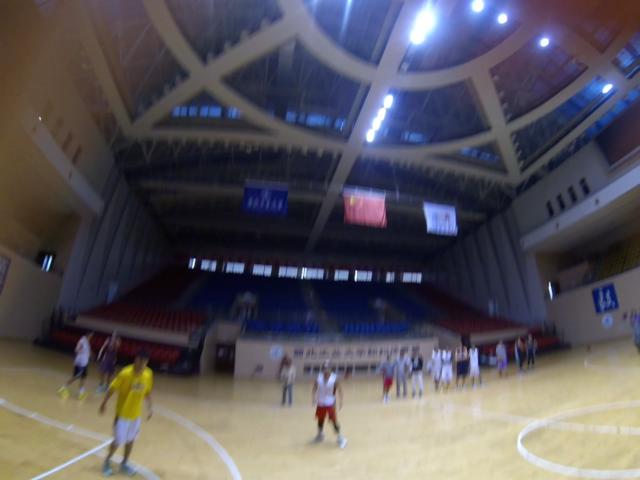}
\myfigurethreecol{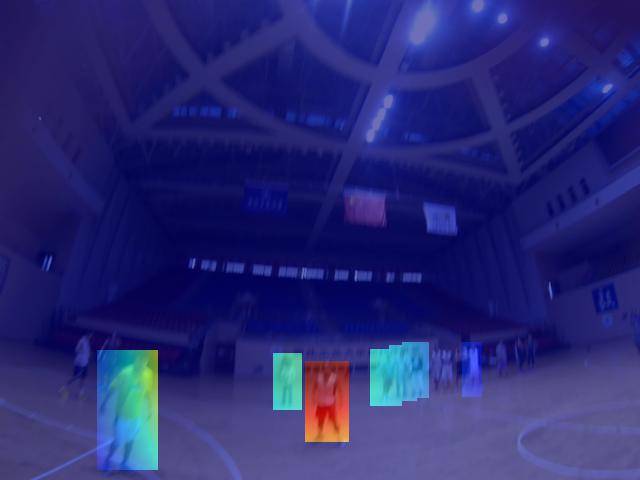}
\myfigurethreecol{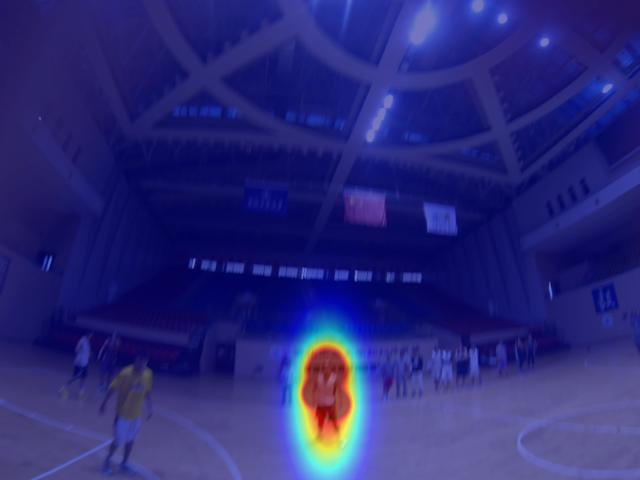}

\myfigurethreecolcaption{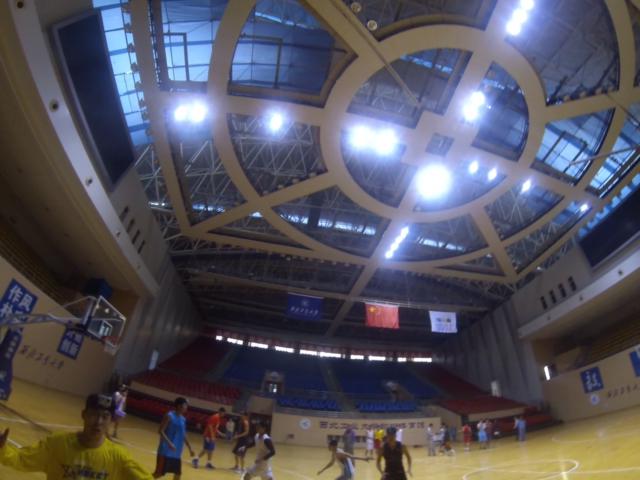}{First-Person RGB}
\myfigurethreecolcaption{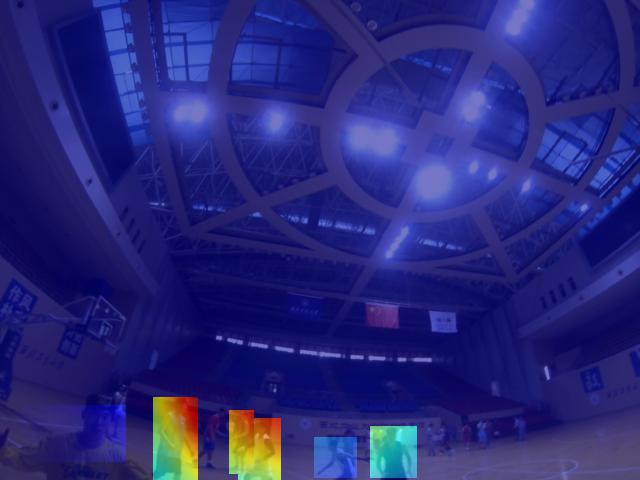}{Pseudo GT}
\myfigurethreecolcaption{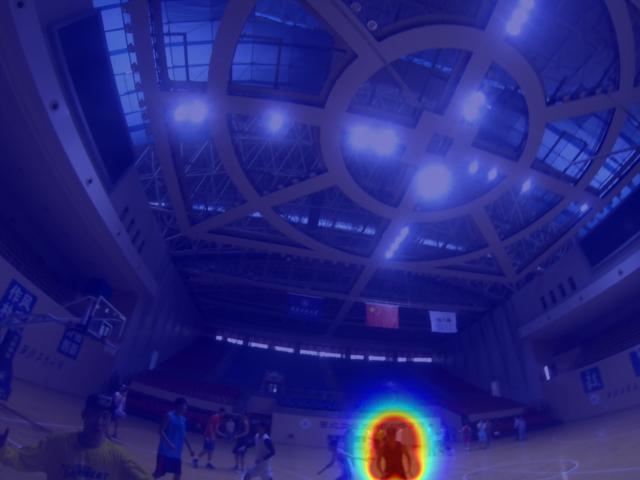}{Ground Truth}

\captionsetup{labelformat=default}
\setcounter{figure}{2}
    \caption{Qualitative comparison of the pseudo ground truth labels obtained via an EgoTransformer versus the actual ground truth. Note that while the pseudo ground truth is not always correct (see the third row), in most cases, it successfully assigns high values around the player with whom the camera wearer will cooperate (see the first two rows). \vspace{-0.5cm}}
    \label{pseudo_gt_fig}
\end{figure}

\captionsetup{labelformat=default}
\captionsetup[figure]{skip=10pt}

For instance,  the location where another person is detected in a first-person image can be used to assess how likely the camera wearer is looking at that person~\cite{Li_2015_CVPR, gberta_2017_RSS}. The size of another person in a first-person image can be used to infer how far the camera wearer is from that person, and hence, how likely will the camera wearer interact with that person (the nearer the more likely). Finally, most person-to-person interactions involve people looking at each other, which imposes a certain pose prior. We can then use such a pose prior to predict whether two people will cooperate with each other in the near future based on whether another person is looking at the camera wearer at present.

\captionsetup{labelformat=empty}
\captionsetup[figure]{skip=5pt}

\begin{figure*}
\centering


\myfiguresixcol{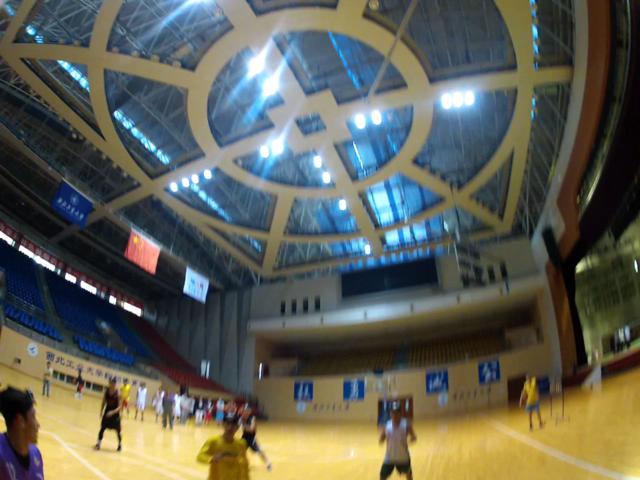}
\myfiguresixcol{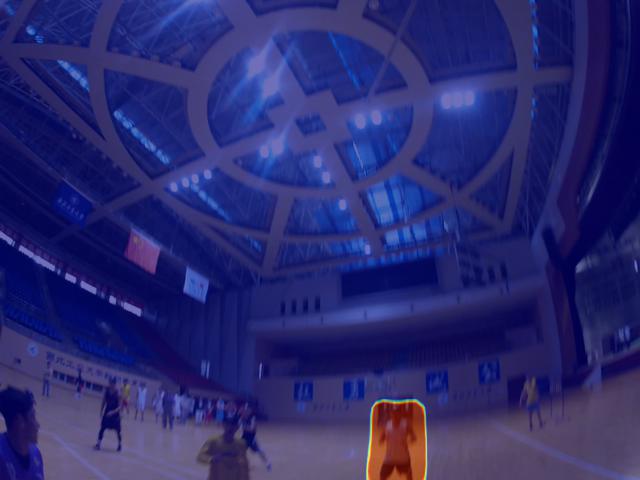}
\myfiguresixcol{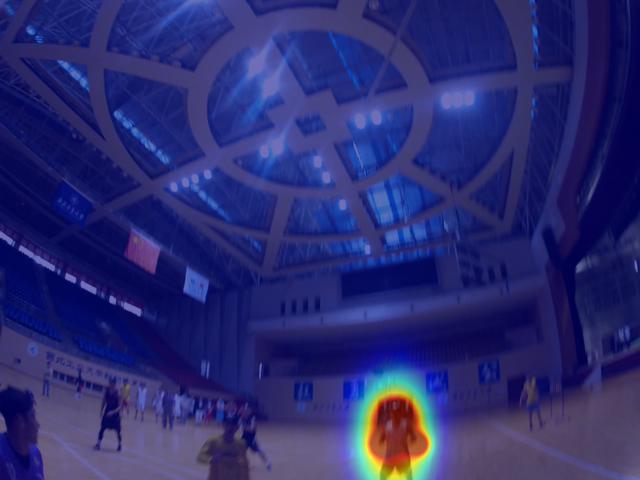}
\myfiguresixcol{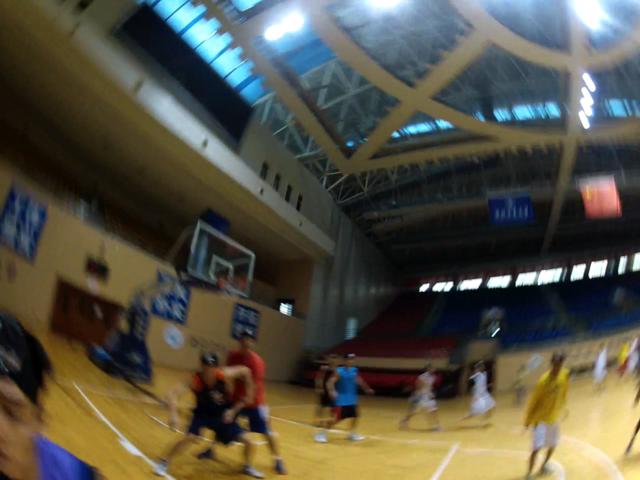}
\myfiguresixcol{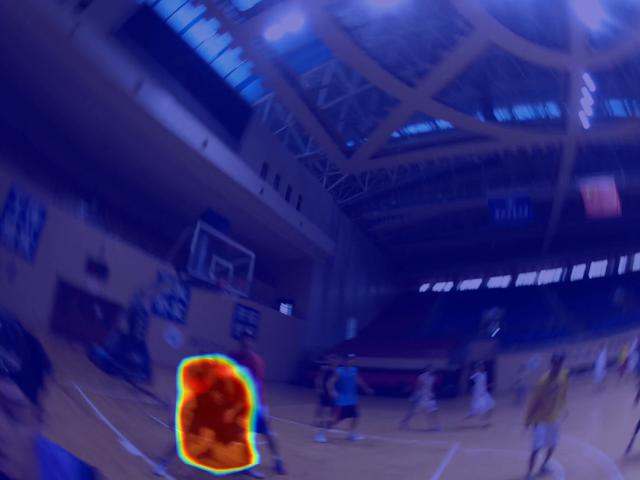}
\myfiguresixcol{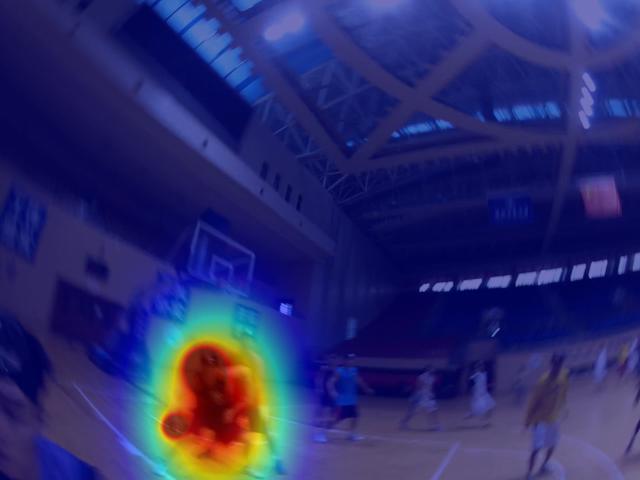}

\myfiguresixcol{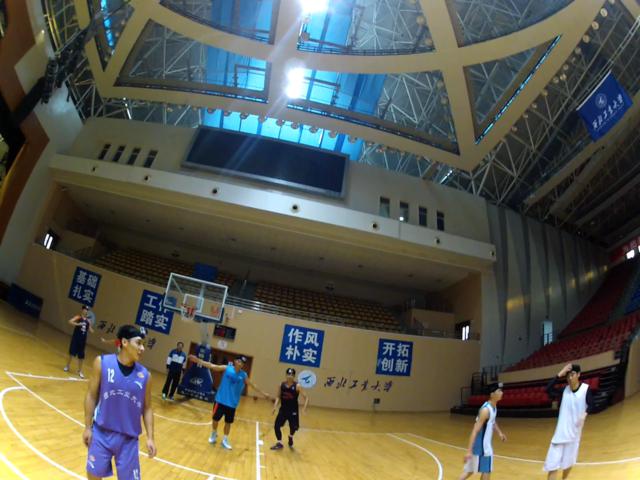}
\myfiguresixcol{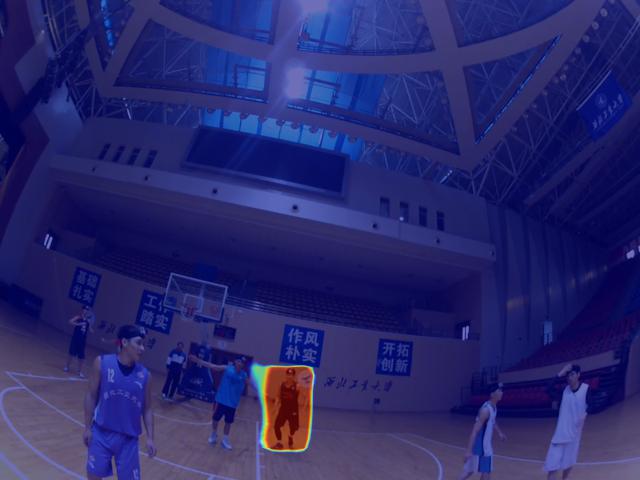}
\myfiguresixcol{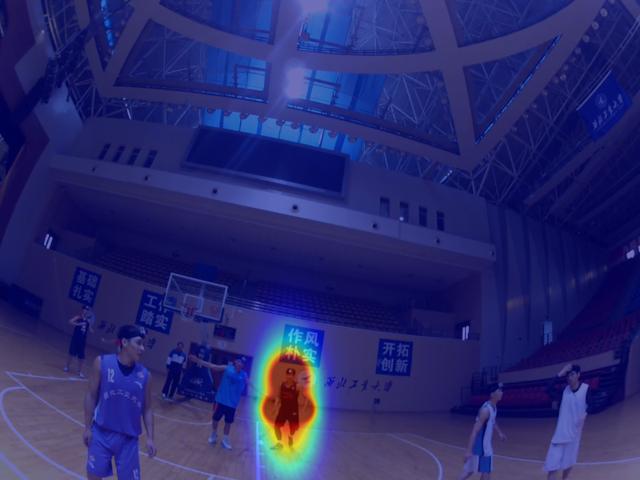}
\myfiguresixcol{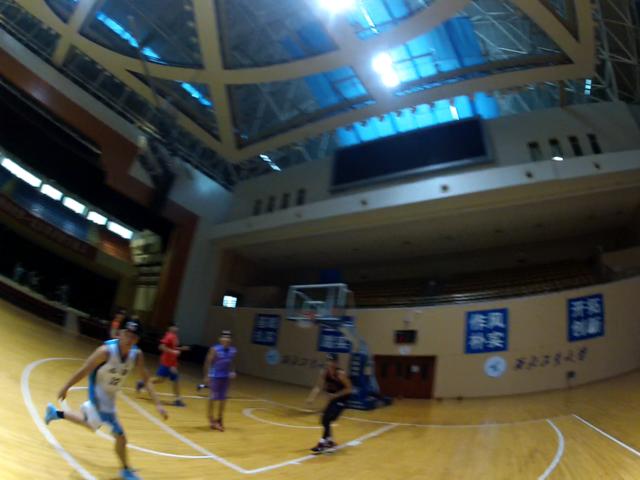}
\myfiguresixcol{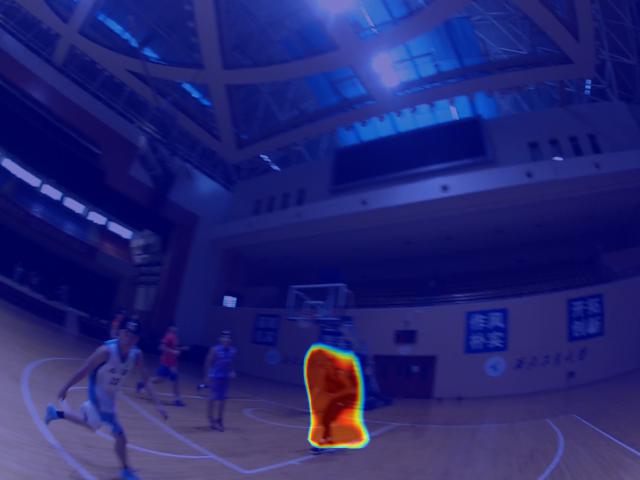}
\myfiguresixcol{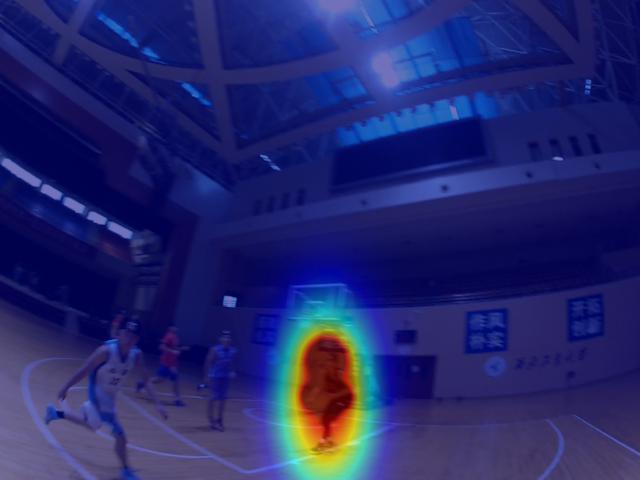}

\myfiguresixcolcaption{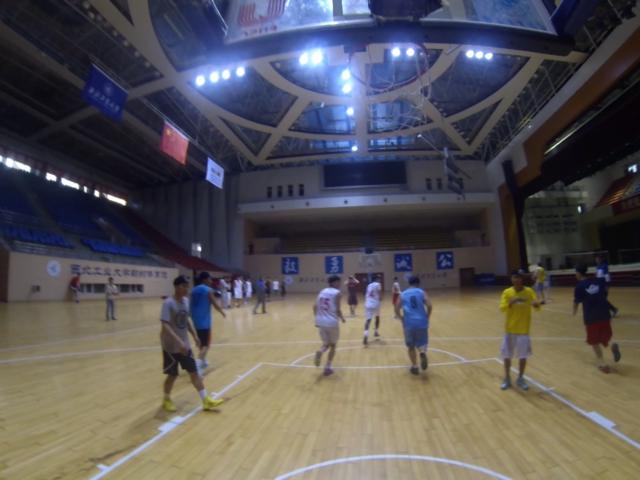}{First-Person RGB}
\myfiguresixcolcaption{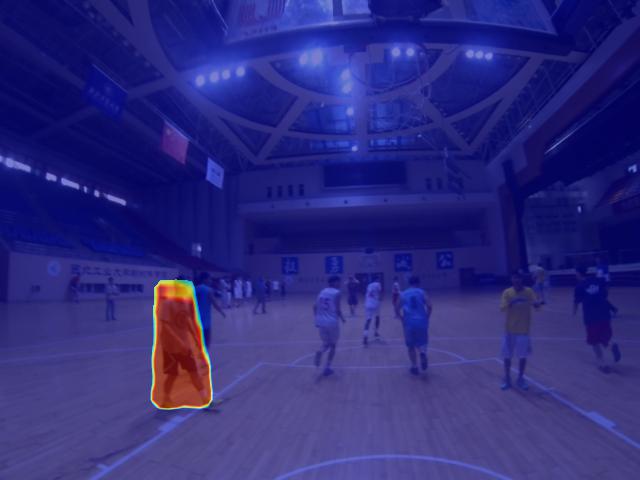}{Our Prediction}
\myfiguresixcolcaption{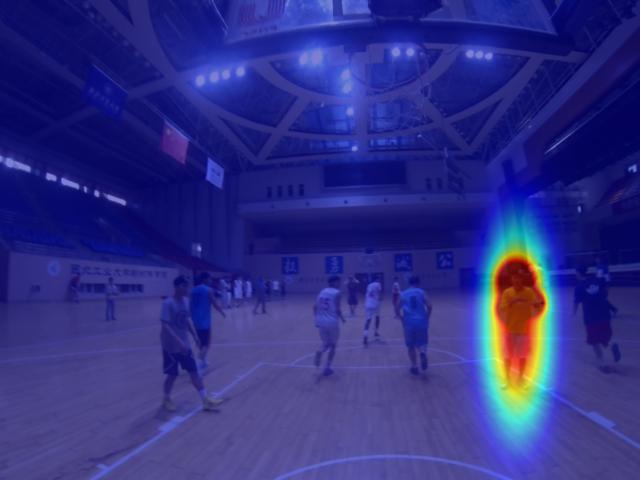}{Ground Truth}
\myfiguresixcolcaption{./paper_figures/qual_results_v2/input/1_GOPR0064_80850.jpg}{First-Person RGB}
\myfiguresixcolcaption{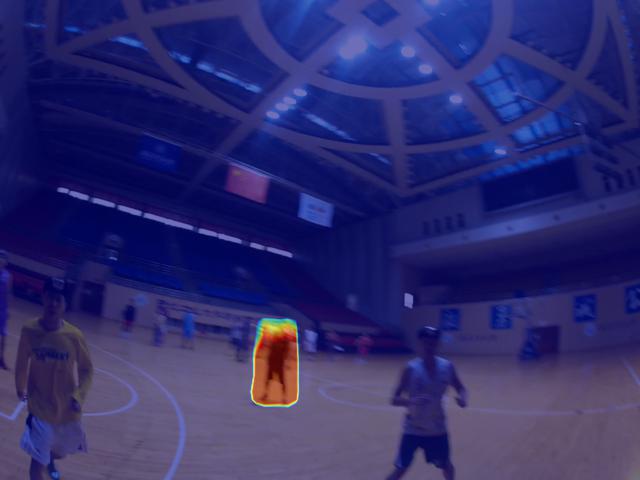}{Our Prediction}
\myfiguresixcolcaption{./paper_figures/qual_results_v2/gt/1_GOPR0064_80850.jpg}{Ground Truth}

\captionsetup{labelformat=default}
\setcounter{figure}{3}
    \caption{The qualitative cooperative basketball intention prediction results. Despite not using any manually annotated first-person labels during training, in most cases, our cross-model EgoSupervision method correctly predicts with whom the camera wearer will cooperate (the first two rows). In the third row, we also illustrate two cases where our method fails to produce  correct predictions. \vspace{-0.5cm}}
    \label{preds_fig}
\end{figure*}

\captionsetup{labelformat=default}
\captionsetup[figure]{skip=10pt}

We express our pseudo ground truth data $\hat{y}$ using these three characteristics using what we refer to as an EgoTransformer scheme:

\vspace{-0.3cm}
\begin{equation}\label{pgt_eq}
\begin{split}
\hat{y} = & \Big[ \sum_{j=1}^n V(B_j, \phi_{size}(B_j)) \cdot V(B_j,\phi_{pose}(B_j))\Big] \cdot \phi_{loc} (B)
\end{split}
\end{equation}

where $n$ denotes the number of detected bounding boxes in a given image, $B_j$ depicts a $j^{th}$ bounding box, $V$ is a function that takes two inputs: 1) a bounding box $B_j$, and 2) a scalar value $v$, and outputs a $H \times W$ dimensional mask by assigning every pixel inside this bounding box $B_j$ to $v$, and zeros to all the pixels outside $B_j$. Here, $H$ and $W$ depict the height and the width of the original input image. Finally, $\phi_{size}(B_j) \in \mathbb{R}^{1 \times 1}$ and $\phi_{pose}(B_j) \in \mathbb{R}^{1 \times 1}$ are scalars that capture the size and pose priors associated with a bounding box $B_j$, while $\phi_{loc} \in \mathbb{R}^{H \times W}$ is a first-person location prior of the same dimensions as the original input image. 

Intuitively, the formulation above operates by first assigning a specific value to each of the detected bounding boxes. This yields a $H \times W$ dimensional prediction map where every pixel that does not belong to any bounding boxes is assigned a zero value. Then, this prediction map is multiplied with the location prior $\phi_{loc} \in \mathbb{R}^{H \times W}$ (using elementwise multiplication). Finally, all the values are normalized to be in range $[0,1]$, which produces our final pseudo ground truth labels. We now explain each of the components in more detail.

\textbf{Egocentric Location Prior.} The location of the camera wearer's visual attention is essential for inferring his/her cooperative intentions.  We know that a first-person camera is aligned with the person's head direction, and thus, it captures exactly what the camera wearer sees. As a result, the way the camera wearer positions himself with respect to other players affects the location where these players will be mapped in a first-person image. 

Instead of assuming any specific location a-priori (e.g. a center prior), as is done in~\cite{Li_2015_CVPR,DBLP:journals/ijcv/LeeG15}, we find the egocentric location prior directly from the data. As before, Let $B \in \mathbb{R}^{n \times 5}$ denote the bounding boxes detected by a pretrained network. Then we can compute  $\phi_{loc} \in \mathbb{R}^{H \times W}$ as follows:


\vspace{-0.3cm}
\begin{equation}
\begin{split}
\phi_{loc}(B)=\sum_{j=1}^n V(B^{(i)}_j,c^{(i)}_j) \cdot \frac{1}{N} \sum_{i=1}^N  \sum_{j=1}^n V(B^{(i)}_j,c^{(i)}_j))\nonumber
\end{split}
\end{equation}

where $c^{(i)}_j$ is the predicted confidence of the $j^{th}$ bounding box in the $i^{th}$ image. Intuitively, the first term $\sum_{j=1}^n V(B_j,c^{(i)}_j)$ depicts a $H \times W$ dimensional mask that is obtained by assigning confidence values to all pixels in their respective bounding boxes in the current image, and zero values to the pixels outside the bounding boxes. The second term $\frac{1}{N} \sum_{i=1}^N  \sum_{j=1}^n V(B_j,c^{(i)}_j))$ also depicts a $H \times W$ dimensional mask that is obtained using this same procedure but across the entire training training dataset rather than a single image. In other words, the second term captures the locations in a first-person image where the bounding box predictions are usually most dense. 

We conjecture, that  $\phi_{loc}(I_i)$ can then be used to approximate the camera wearer's visual attention location, which is essential for inferring the camera wearer's cooperative intentions.


\captionsetup{labelformat=empty}
\captionsetup[figure]{skip=5pt}

\begin{figure*}[t]
\centering

\myfiguresixcol{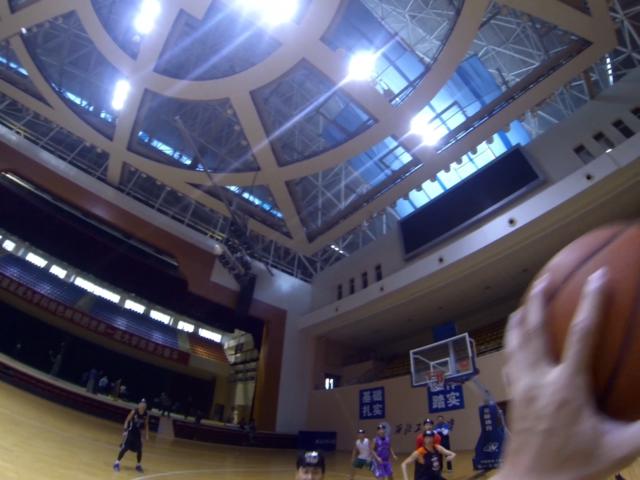}
\myfiguresixcol{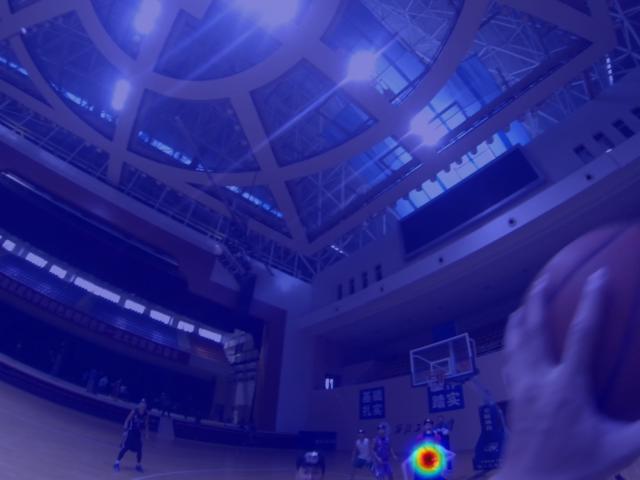}
\myfiguresixcol{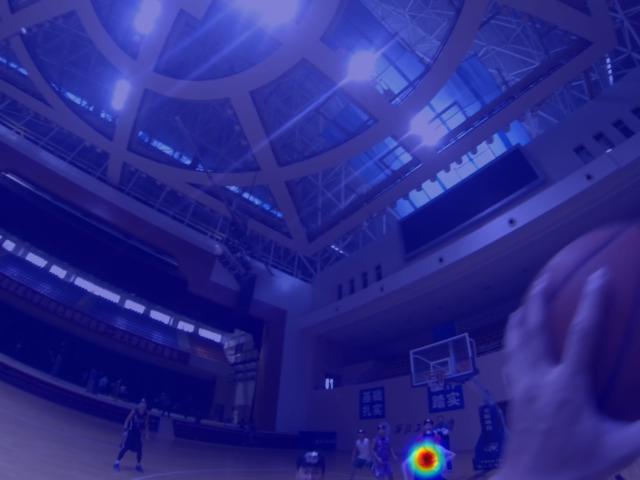}
\myfiguresixcol{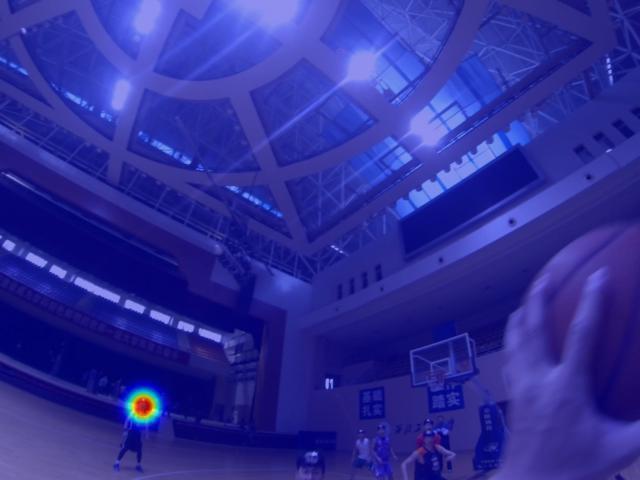}
\myfiguresixcol{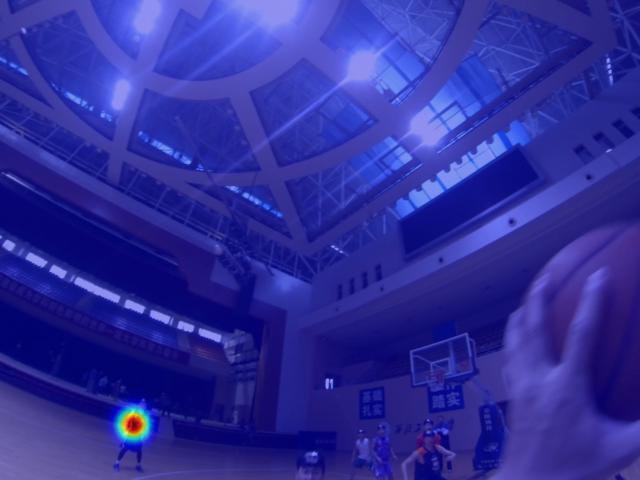}
\myfiguresixcol{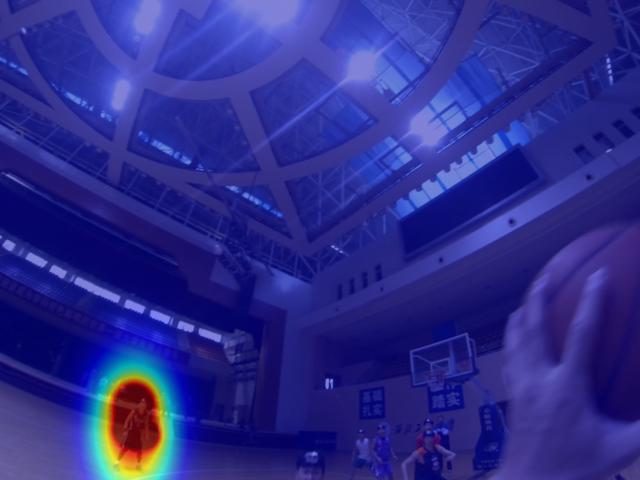}

\myfiguresixcolcaption{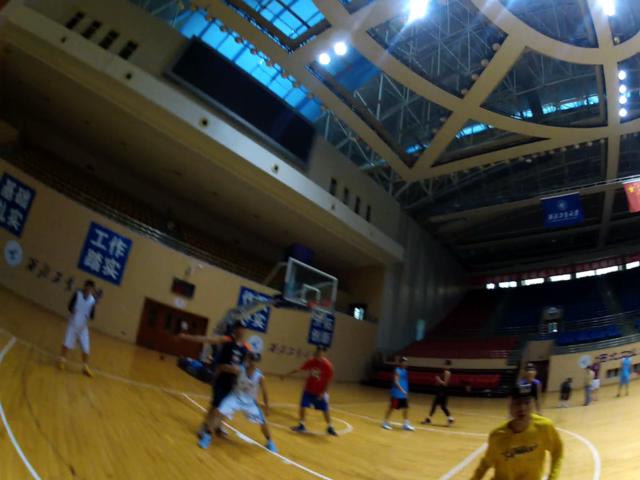}{First-Person RGB}
\myfiguresixcolcaption{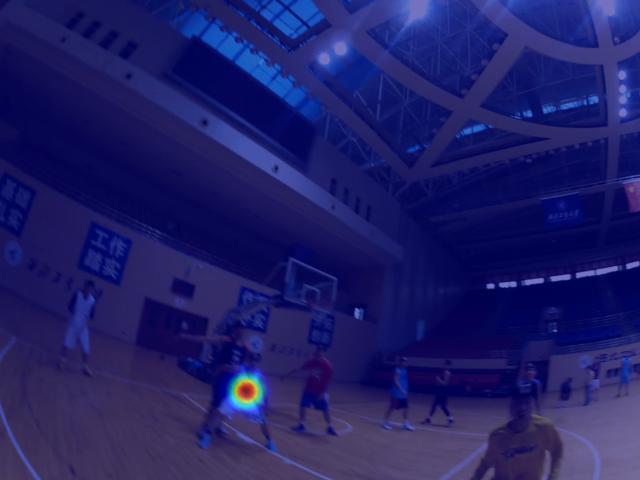}{Subject-1}
\myfiguresixcolcaption{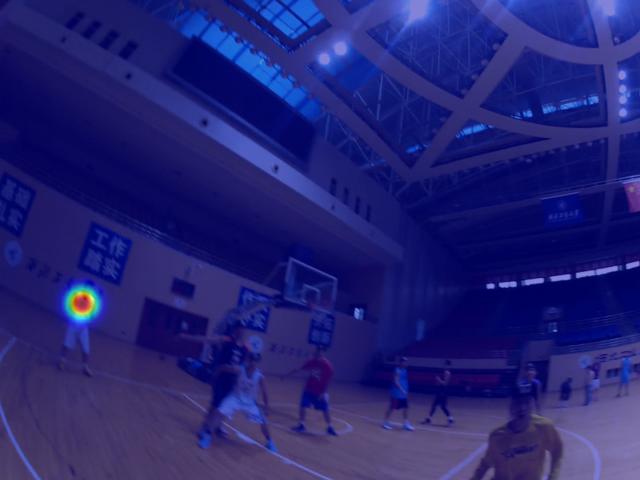}{Subject-2}
\myfiguresixcolcaption{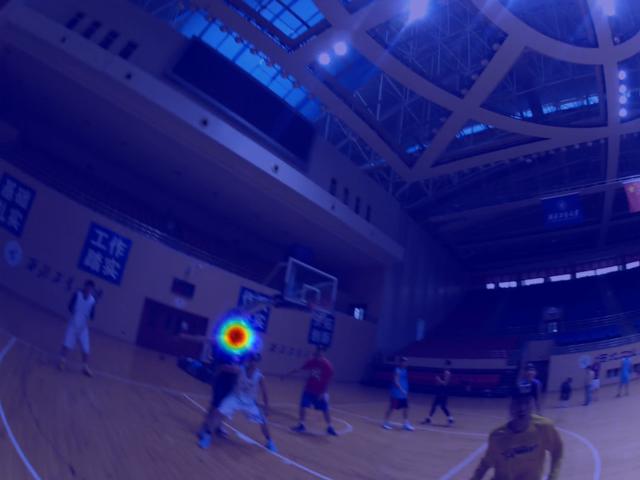}{Subject-3}
\myfiguresixcolcaption{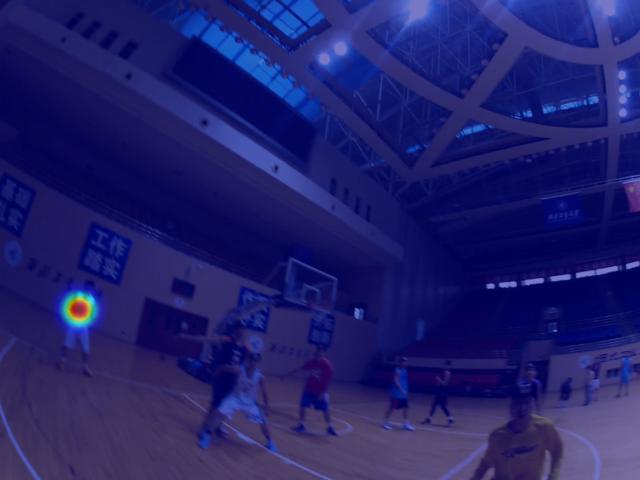}{Subject-5}
\myfiguresixcolcaption{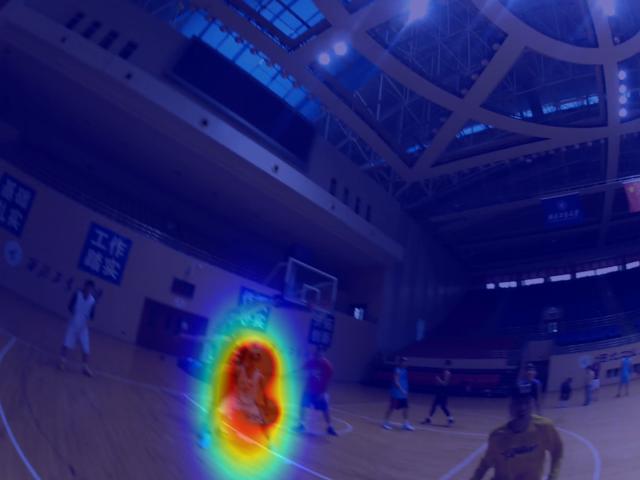}{Ground Truth}

\captionsetup{labelformat=default}
\setcounter{figure}{4}
\caption{Several qualitative examples from the top $4$ performing subjects in our conducted human study. Each subject specified their prediction by clicking on the person, with whom he/she thought the camera wearer was going to cooperate. We then placed a fixed size Gaussian around the location of the click. Note that based on these results, we can conclude that some instances of this task are quite difficult even for humans, i.e. in these examples, there is no general consensus among the subjects' responses. \vspace{-0.4cm}}

    \label{human_preds}
\end{figure*}

\captionsetup{labelformat=default}
\captionsetup[figure]{skip=10pt}

\textbf{Egocentric Size Prior.} Spatial $3D$ cues provides important information to infer the camera wearer's intentions~\cite{park_ego_future,park_cvpr:2017}. For instance, the camera wearer is more likely to cooperate with a player who is near him/her. We propose to capture this intuition, by exploiting an egocentric size prior. We know that the size of a bounding box in a first-person image is inversely related to the distance between the camera wearer and the person in the bounding box. Thus, let $h_j$ be the height of the bounding box $B_j$. Then we express the egocentric size prior $\phi_{size}(B_j) \in \mathbb{R}^{1 \times 1}$ for a given bounding box as:

\vspace{-0.3cm}
\begin{equation}
\begin{split}
\phi_{size}(B_j)=   \exp{(-\frac{\sigma}{h_j})}\nonumber
\end{split}
\end{equation}

where $\sigma$ denotes a hyperparameter controlling how much to penalize small bounding boxes. Such a formulation allows us to capture the intuition that the camera wearer is more likely to cooperate with players who are physically closer to him/her.

\textbf{Egocentric Pose Prior.}  In basketball, people tend to look at each other to express their intentions before actually performing cooperative actions. Detecting whether a particular person is facing the camera wearer can be easily done by examining the $x$ coordinates  of the paired body parts such as eyes, arms, legs, etc of a person detected in a first-person image. For instance, if a particular person is facing the camera wearer then, we will observe that for most of his/her paired parts visible in a first-person image the following will be true: $x(right\_part)<x(left\_part)$. In other words, the right parts of that person's body will have smaller $x$ coordinate value in a first-person image, than the left parts. We use this intuition to encode the egocentric pose prior  $\phi_{pose}(B_j) \in \mathbb{R}^{1 \times 1}$ for a given bounding box $B_j$ as follows:

\vspace{-0.3cm}
\begin{equation}
\begin{split}
\phi_{pose}(B_j)=\frac{1}{|\mathcal{P}|}\sum_{p \in \mathcal{P}} 1 \{x(right\_part)<x(left\_part) \} \nonumber
\end{split}
\end{equation}

where $\mathcal{P}$ is the set of all paired parts, and $1 \{x(right\_part)<x(left\_part) \}$ is an indicator function that returns $1$ if the $x$ coordinate of the right part in a first-person image is smaller than the $x$ coordinate of the left part. The computed value $\phi_{pose}(B_j)$ can then be viewed as a confidence that a person in the bounding box $B_j$ is facing the camera wearer, which is an important cue for inferring the camera wearer's cooperative intentions.

\textbf{Pseudo Ground Truth.} We then combine all the above discussed components into a unified framework using the Equation~\ref{pgt_eq}. Such a formulation allows us to automatically construct pseudo ground truth labels from the outputs of a pretrained multi-person pose estimation network. We illustrate several examples of our obtained pseudo ground truth labels in Figure~\ref{pseudo_gt_fig}. Notice that while our computed pseudo ground truth is not always correct, in many cases it correctly captures the player with whom the camera wearer will cooperate in the near future. In our experimental section, we will demonstrate that despite the imperfections of our pseudo ground truth labels, we can use them to obtain a model that is almost as good as the model trained in a fully supervised fashion using manually annotated cooperation labels.

\subsection{Cross-Model EgoSupervision}

After obtaining the pseudo ground truth data $\hat{y}$, we train our cooperative basketball intention FCN using the cross-model EgoSupervision scheme as shown in Figure~\ref{fig:train_arch}. We employ a multi-person pose estimation network from~\cite{DBLP:journals/corr/CaoSWS16} as our base model, which is used to predict the 1) pose estimates of all people in a given image and 2) their bounding boxes. The parameters inside the base network are fixed throughout the entire training procedure. At each iteration, the outputs from the base network are fed to the EgoTransformer, which transforms them into the pseudo ground truth cooperate intention labels. These pseudo ground truth labels are then used as a supervisory signal to train our cooperative basketball intention FCN using a sigmoid cross entropy per-pixel loss as illustrated in Equation~\ref{CE_loss_eq}.

\subsection{Implementation Details}

For all of our experiments, we used a Caffe deep learning library~\cite{jia2014caffe}. As our base FCN model we used a multi-person pose estimation network from~\cite{DBLP:journals/corr/CaoSWS16}. Inspired by the success of this method, we also used the same architecture for our cooperative basketball intention FCN. During training, we optimized the network for $5000$ iterations with a learning rate of $10^{-7}$, the momentum equal to $0.9$, the weight decay of $0.0005$, and the batch size of $15$. The weights inside the base FCN network were fixed throughout the entire training procedure. To compute the egocentric size prior mask we used $\sigma = 10$.

%

\section{Cooperative Basketball Intention Dataset}
\label{data_sec}

We build upon the dataset from~\cite{DBLP:journals/corr/BertasiusYPS16}, which captures first-person basketball videos of $48$ distinct college-level players in an unscripted basketball game. The work in~\cite{DBLP:journals/corr/BertasiusYPS16} studies a basketball performance assessment problem, and provides $401$ training and $343$ testing examples of basketball cooperations among players from $10.3$ hours of videos.

To obtain ground truth labels corresponding to the specific players, with whom the camera wearer cooperated, we look at the video segments corresponding to all such cooperation. We then identify the player with whom the camera wearer cooperated, go back to the frame about $2$ seconds before the cooperation happens, and label that player with a bounding box. The ground truth data is then generated by placing a Gaussian inside the bounding box, according to the height and width of the bounding box. 

Once again we note that these labels are only used for the evaluation purposes, and also to train other baseline models. In comparison, our method learns to detect the players with whom the camera wearer will cooperate, without relying on manually annotated intention labels.



\section{Experimental Results}

In this section, we present quantitative and qualitative results for our cooperative basketball intention prediction task. To compute the accuracy of each method, we select the player in the image with the maximum predicted probability as the the final prediction and then compute the fraction of all the correct predictions across the entire testing dataset. 

    \setlength{\tabcolsep}{3pt}
   
         \begin{table}
    \begin{center}
    \begin{tabular}{  c | c |}
    \hline
    \multicolumn{1}{| c |}{\em Human Subjects} & {\em Accuracy}\\ \hline
    \multicolumn{1}{| c |}{Subject-4} & 0.802\\ 
     \multicolumn{1}{| c |}{Subject-2} & 0.895\\ 
      \multicolumn{1}{| c |}{Subject-3} & 0.901\\ 
        \multicolumn{1}{| c |}{Subject-5} & 0.904\\ 
    \multicolumn{1}{| c |}{Subject-1} & \bf 0.927\\ \hline
  
    \end{tabular}
    \end{center}\vspace{-.3cm}
     \caption{Quantitative human study results on our cooperative basketball intention task. We ask $5$ subjects to predict a player in the first-person image, with whom they think the camera wearer will cooperate after $2$ seconds. We then compute the accuracy as the fraction of correct responses. The results indicate that most subjects achieve the accuracy of about $90\%$. We conjecture that Subject-4 may be less familiar with the basketball game thus, the lower accuracy. \vspace{-0.2cm}}
    \label{human_study_table}
   \end{table}

\subsection{Human Study}
\label{human_study_sec}

First, to see how well humans can predict cooperative basketball intention from first-person images, we conduct a human study consisting of $5$ human subjects. Each subject is shown $343$ testing images one at a time, and asked to click on the player in an image, with whom he/she thinks the camera wearer will cooperate $2$ seconds from now. Then the accuracy of each subject is evaluated as the fraction of correct responses.  

We present these results in Table~\ref{human_study_table}, and demonstrate that this task is not trivial even for humans: most of the subjects achieve about $90\%$ accuracy on our task, which is solid but not perfect. We also point out that we did not collect information on how familiar each subject was with basketball. However, based on the results, we conjecture that Subject-4 who achieved almost $10\%$ lower accuracy than the other subjects was probably not very familiar with basketball, which contributed to his lower performance. In Figure~\ref{human_preds}, we also visualize the qualitative examples that human subjects found the most difficult, i.e. in these instances, the predictions among the subjects differed substantially.

    \setlength{\tabcolsep}{3pt}
   
         \begin{table}
    \begin{center}
    \begin{tabular}{  c | F{2cm} |}
    \hline
    \multicolumn{1}{| c |}{\em Method} & {\em Accuracy}\\ \hline
    \multicolumn{1}{| c |}{DCL~\cite{LiYu16}} & 0.222\\ 
    \multicolumn{1}{| c |}{MPP-pretrained~\cite{DBLP:journals/corr/CaoSWS16}} & 0.586\\  
     \multicolumn{1}{| c |}{DeepLab$^{\ddagger}$~\cite{DBLP:journals/corr/ChenYWXY15}} & 0.644\\    
      \multicolumn{1}{| c |}{Pseudo GT} & 0.665\\   
      \multicolumn{1}{| c |}{ResNet-50$^{\ddagger}$~\cite{He2015}} & 0.675\\ 
       \multicolumn{1}{| c |}{PSPNet$^{\ddagger}$~\cite{DBLP:journals/corr/ZhaoSQWJ16}} & 0.695\\ 
       \multicolumn{1}{| c |}{ResNet-101$^{\ddagger}$~\cite{He2015}} & 0.706\\ 
       \multicolumn{1}{| c |}{DeepLab-v2$^{\ddagger}$~\cite{CP2016Deeplab}} & 0.757\\        
       \multicolumn{1}{| c |}{MPP-finetuned$^{\ddagger}$~\cite{DBLP:journals/corr/CaoSWS16}} & \bf 0.778\\ \hline
       \multicolumn{1}{| c |}{\bf CMES} & 0.775\\ \hline 
       
    \end{tabular}
    \end{center}\vspace{-.3cm}
     \caption{The quantitative cooperative basketball intention results evaluated as the fraction of correct predictions. We compare our Cross-Model EgoSupervision (CMES) scheme with a variety of supervised methods (marked by $\ddagger$). These results indicate that even without using manually annotated intention labels, our method outperforms most supervised methods, and produces almost identical performance as our main baseline ``MPP-finetuned''.\vspace{-0.2cm}}
    \label{cbi_results_table}
   \end{table}


\subsection{Quantitative Results}

In Table~\ref{cbi_results_table}, we present quantitative cooperative basketball intention results of our method and several other baselines. As our baselines, we use a collection of methods that were successfully used for other computer vision tasks such as image classification, semantic segmentation or saliency detection. These include a 1) Deep Contrast Saliency (DCL) method~\cite{LiYu16}, 2-3) several variations of highly successful DeepLab semantic segmentation systems~\cite{DBLP:journals/corr/ChenYWXY15,CP2016Deeplab} adapted to our task, 4-5) image classification ResNets~\cite{He2015} adapted to our task, 6) one of the top performing semantic segmentation systems PSPNet~\cite{DBLP:journals/corr/ZhaoSQWJ16}, 7-8) a pretrained and finetuned multi-person pose estimation (MPP) network~\cite{DBLP:journals/corr/CaoSWS16}, and 9) a pseudo ground truth  obtained from our EgoTransformer.

Note that our Cross-Model EgoSupervision (CMES) method is based on an MPP network architecture~\cite{DBLP:journals/corr/CaoSWS16}, and thus, as our main baseline we use the  ``MPP-finetuned'' method, which uses the manually labeled bounding box intention labels to infer with whom the camera wearer will interact. In contrast to this baseline, our CMES method is only trained on the automatically generated pseudo ground truth labels. We note that the supervised methods employing manually labeled data are marked with $^{\ddagger}$. We now discuss several interesting observations based on these results.

 \setlength{\tabcolsep}{3pt}
   
         \begin{table}
    \begin{center}
    \begin{tabular}{  c | F{2.5cm} | F{2.5cm} |}
    \cline{2-3}
    & \multicolumn{2}{ c |}{{\em Accuracy}}\\
    \hline
    \multicolumn{1}{| c |}{\em Method} & {\em pseudo GT} & {\em Trained Model}\\ \hline

    \multicolumn{1}{| c |}{no $\phi_{loc}$} & 0.481 & 0.560\\   
    \multicolumn{1}{| c |}{no $\phi_{pose}$} & 0.557 & 0.694\\ 
     \multicolumn{1}{| c |}{no $\phi_{size}$} & 0.571 & 0.731\\ \hline
        \multicolumn{1}{| c |}{\bf Ours-Full} & \bf 0.665 & \bf 0.775\\ \hline
    \end{tabular}
    \end{center}\vspace{-.3cm}
     \caption{The quantitative ablation studies documenting the importance of each component in our EgoTransformer scheme. We separately remove each of $\phi_{loc}$, $\phi_{size}$, $\phi_{pose}$ and investigate how the accuracy changes. The second column in the table denotes the accuracy of a pseudo ground truth, while the third column depicts the accuracy of our trained model. Based on these results, we can conclude that each component of our EgoTransformer is essential for an accurate cooperative basketball intention prediction. \vspace{-0.5cm}}
    \label{egotransformer_results_table}
   \end{table}



\textbf{Comparison with the Supervised Methods.} Based on the results, we observe that despite not using manually annotated bounding box intention labels, our method outperforms a number of supervised baselines and achieves almost equivalent results to our main baseline ``MPP-finetuned'', which  was trained using manually annotated cooperative intention labels. Thus, these results indicatee the effectiveness of our cross-model EgoSupervision scheme.

\textbf{Comparison with the Pseudo Ground Truth.} One interesting and a bit surprising observation from Table~\ref{cbi_results_table}, is that our cross-model EgoSupervision model achieves substantially better accuracy than the pseudo ground truth, which was used to optimize our model. We conjecture that this happens due to the following reasons. The pseudo ground truth labels are constructed using three different signals: 1) an egocentric location prior, 2) an egocentric size prior, and 3) an egocentric pose prior.  Note, that our constructed pseudo ground truth does not incorporate any visual appearance information, i.e. it does not consider how the players look like. In contrast, during training, our network, learns what are the visual appearance cues indicative of the players with high pseudo ground truth values. Arguably, such visual cues provide a stronger signal for a cooperative intention recognition, which then leads to a substantially better performance than the pseudo ground truth labels.

\subsection{Qualitative Results}


In Figure~\ref{preds_fig}, we present our qualitative results, where we show that in most cases, our model successfully learns to predict with whom the camera wearer will cooperate. Furthermore, to gain a better understanding of what the network learned, in Figure~\ref{filters_fig}, we visualize the activations inside the second to last FCN's layer. Note that our network has high activation values around the faces of people with whom the camera wearer intends to cooperate. This makes intuitive sense, as face is probably the most useful cue to recognize the camera wearer's intention to cooperate.



\subsection{Ablation Experiments}

In Table~\ref{egotransformer_results_table}, we present the results analyzing the behavior of our EgoTransformer scheme. Earlier we discussed that to implement our EgoTransformer scheme we exploit three characteristics: 1) egocentric location prior $\phi_{loc}$ , 2) egocentric size prior $\phi_{size}$ , and 3) egocentric pose prior $\phi_{pose}$. We want to investigate how much each of these priors affect 1) the quality of our generated pseudo ground truth data, and 2) the quality of our model trained using such pseudo ground truth. To do this, we run experiments with three baselines where for each baseline we remove one of  $\phi_{loc}, \phi_{size},$ or $\phi_{pose}$ components. We denote these three baselines as ``no $\phi_{loc}$'', ``no $\phi_{size}$'' and ``no  $\phi_{pose}$'' respectively. Finally, we include the results of our model using the full EgoTransformer scheme.

\captionsetup{labelformat=empty}
\captionsetup[figure]{skip=5pt}

\begin{figure}
\centering

\myfigurethreecol{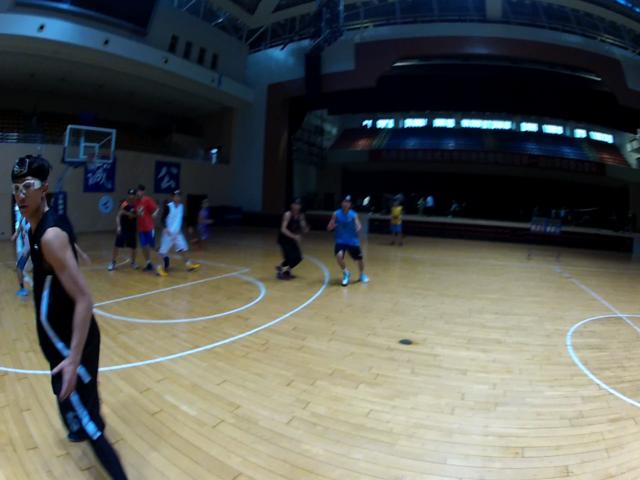}
\myfigurethreecol{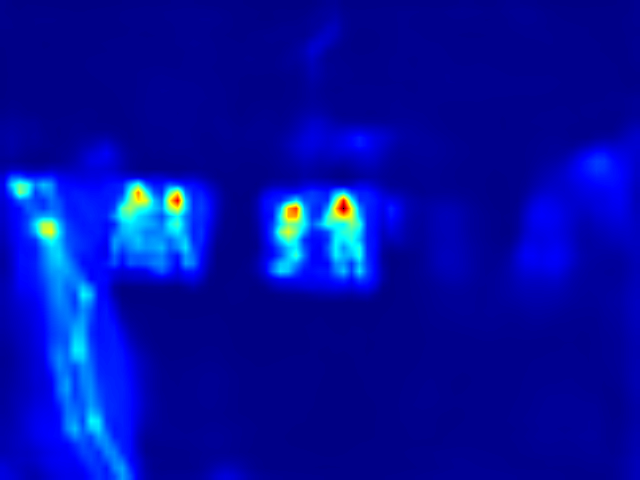}
\myfigurethreecol{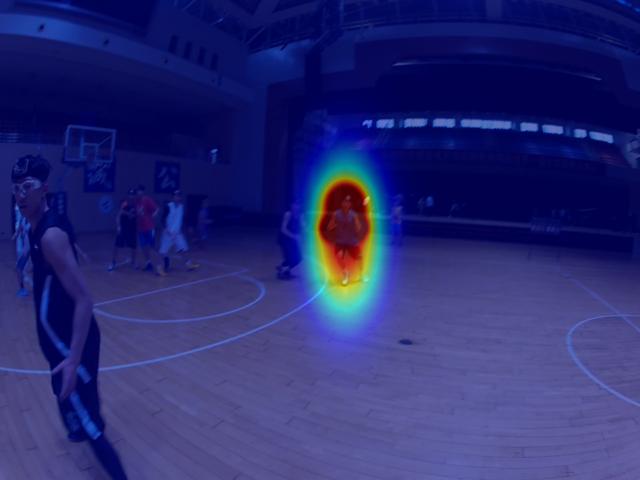}

\myfigurethreecol{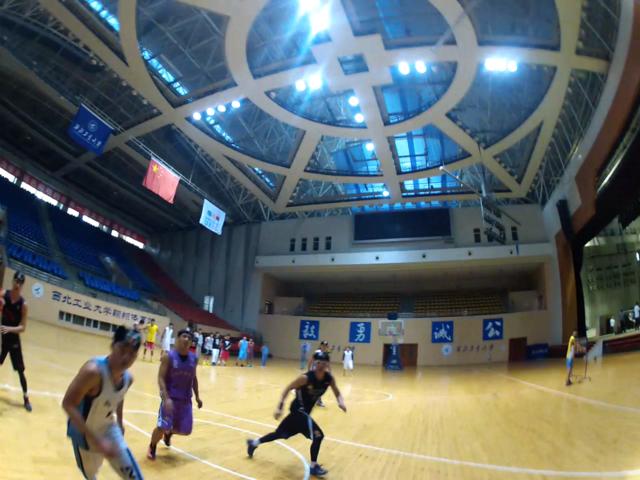}
\myfigurethreecol{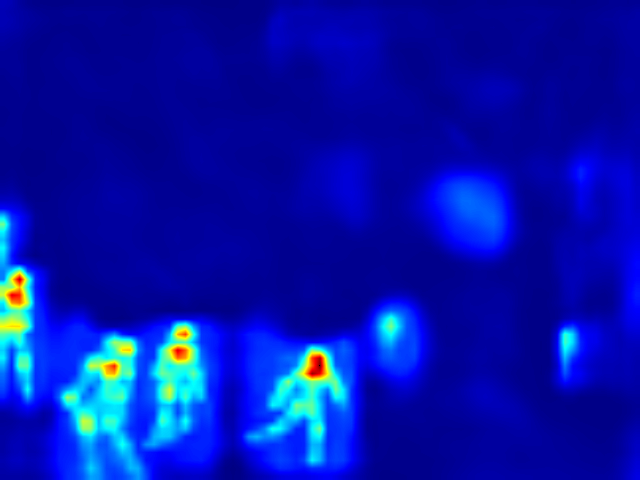}
\myfigurethreecol{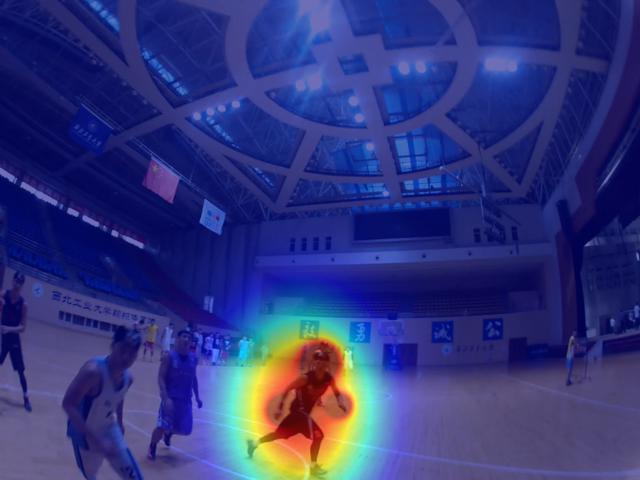}

\myfigurethreecolcaption{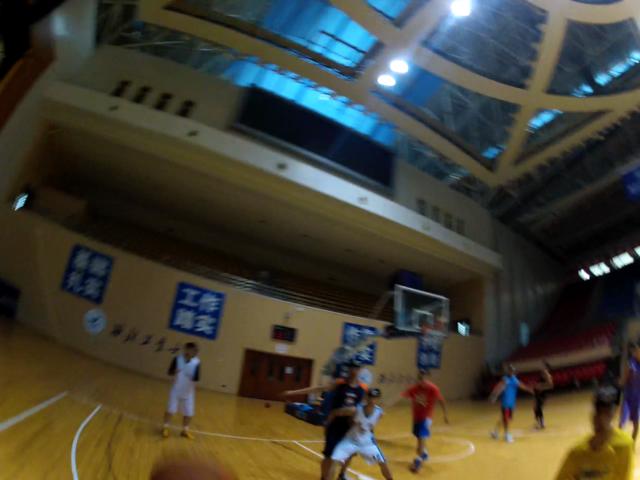}{First-Person RGB}
\myfigurethreecolcaption{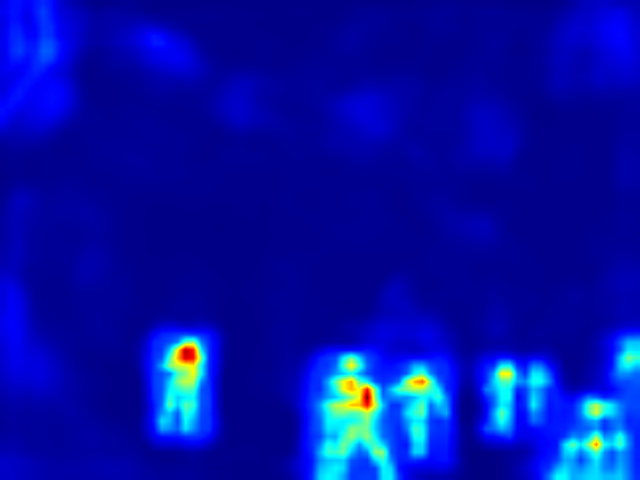}{FCN Activations}
\myfigurethreecolcaption{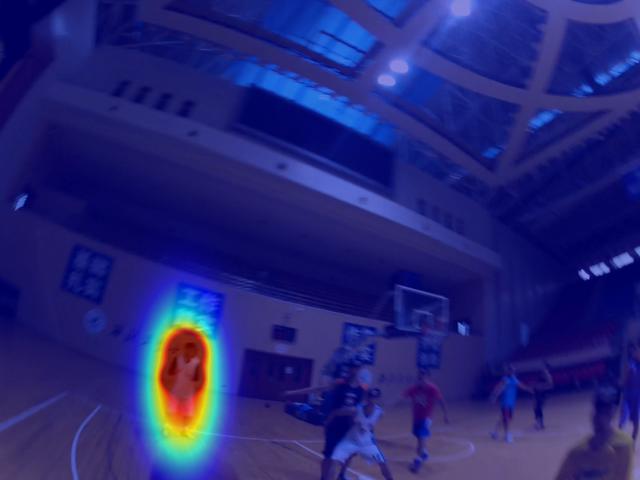}{Ground Truth}

\captionsetup{labelformat=default}
\setcounter{figure}{5}
    \caption{The visualization of the activation values inside the second to last layer in our trained network. Note that the network produces high activation values around the faces of the players in the camera wearer's field of view. This makes intuitive sense, as facial expressions provide the most informative cues for a cooperative basketball intention task. \vspace{-0.5cm}}
    \label{filters_fig}
\end{figure}

\captionsetup{labelformat=default}
\captionsetup[figure]{skip=10pt}

Based on the results in Table~\ref{egotransformer_results_table}, we first observe that each of these components have a significant impact to the quality of pseudo ground truth that we obtain. Specifically, using our full model yields $9.4\%$ better pseudo ground truth results than the second best baseline. Additionally, note that the network trained to the pseudo ground truth of our full model achieves $4.4\%$ higher accuracy than the second best baseline. These results indicate that each component in our EgoTransformer scheme is crucial for learning a high quality cooperative intention model.


\section{Conclusions}



In this work, we present a new task of predicting cooperative basketball intention from a single first-person image. We demonstrate that a first-person image provides strong cues to infer the camera wearer's intentions based on what he/she sees. We use this observation to design a new cross-model EgoSupervision learning scheme that allows us to predict with whom the camera wearer will cooperate, without using manually labeled intention labels.  We demonstrate that despite not using such labels, our method achieves similar or even better results than fully supervised methods.

We believe that our proposed cross-model EgoSupervision scheme could be applied on various other first-person vision tasks without the need to manually collect labels for each of such tasks. In the long run, a learning scheme such as ours could effectively replace the supervised methods, which require costly and time consuming annotation process.



\bibliographystyle{plain}
\footnotesize{
\bibliography{gb_bibliography_v2,bib_hs_v2}}

\end{document}